\documentclass[letterpaper, 10 pt, conference]{ieeeconf} \IEEEoverridecommandlockouts                              

\usepackage{cite}
\usepackage{amsmath,amssymb,amsfonts}
\usepackage{algorithmic}
\usepackage{graphicx}
\usepackage{textcomp}
\usepackage{url}
\usepackage{xcolor}
\def\BibTeX{{\rm B\kern-.05em{\sc i\kern-.025em b}\kern-.08em
    T\kern-.1667em\lower.7ex\hbox{E}\kern-.125emX}}
\begin{document}

\title{\LARGE \bf Painted Heart Beats}

\author{Angshu Adhya$^1$, Cindy Yang$^2$, Emily Wu$^3$, Rishad Hasan$^3$, Abhishek Narula$^3$, Patrícia Alves-Oliveira$^3$\thanks{$^1$~Department of Computer Science, University of Michigan, $^2$~Department of Electrical Engineering, University of Michigan, $^3$~Department of Robotics, University of Michigan $^*$The initial 4 authors are organized alphabetically and contributed equally to this work.}}


\maketitle
\thispagestyle{empty}
\pagestyle{empty}

\begin{abstract}
In this work we present AURA, a framework for synergetic human-artist painting. We developed a robot arm that collaboratively paints with a human artist. The robot has an awareness of the artist's heartbeat through the EmotiBit sensor, which provides the arousal levels of the painter. Given the heartbeat detected, the robot decides to increase proximity to the artist's workspace or retract. If a higher heartbeat is detected, which is associated with increased arousal in human artists, the robot will move away from that area of the canvas. If the artist's heart rate is detected as neutral, indicating the human artist's baseline state, the robot will continue its painting actions across the entire canvas. We also demonstrate and propose alternative robot-artist interactions using natural language and physical touch. This work combines the biometrics of a human artist to inform fluent artistic interactions. 
\end{abstract}


\section{Introduction and Related Work}
In this proposal, we introduce AURA --- a human-robot collaboration framework designed to enhance the creative synergy between artists and robots. AURA builds on previous robot-artist systems, such as FRIDA~\cite{frida} and CoFRIDA~\cite{schaldenbrand2024cofrida}. Taking these two systems, we incorporated multimodal input, including biometric data, voice commands, and real-time visual feedback. Additionally, AURA enables continuous interaction between the human artist and the robot, emphasizing the fluidity of collaboration. Our team combines robotics, generative artificial intelligence (AI), and human-centered design to explore the impact of these expanded capabilities on human-robot interaction and collaboration in the context of arts. This is the first step within a project that aims to understand more about human-robot artistic collaborations. 

\subsection{Wearables within Arts}

\begin{figure}
    \centering
    \includegraphics[width=\columnwidth]{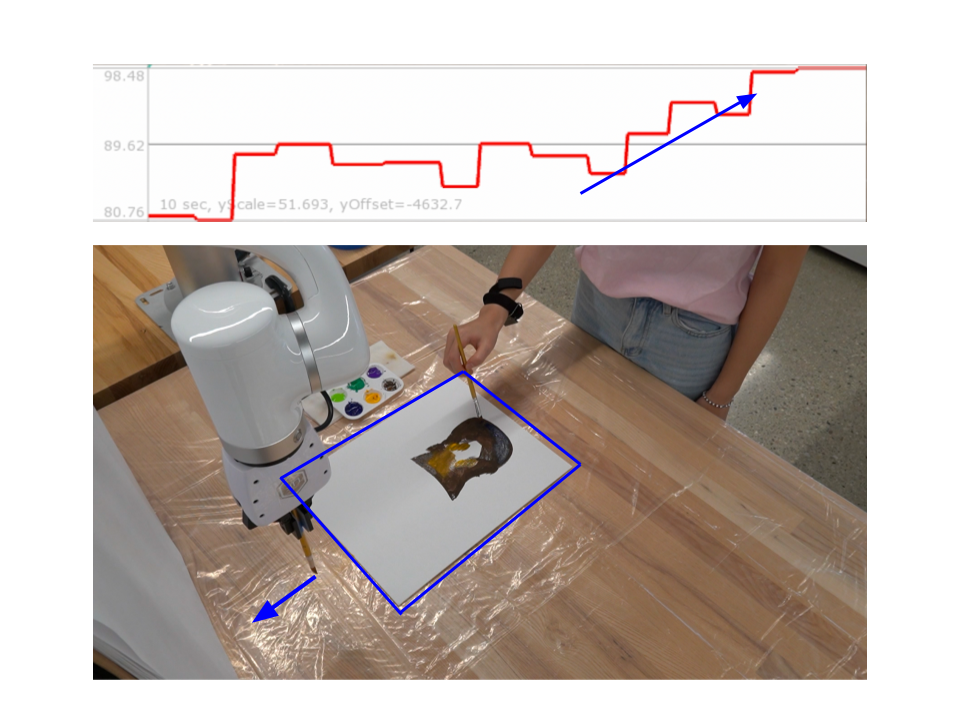}
    \caption{Robot reaction to increased heart rate. When the robot detects increased user heart rate (as shown in the graph), the robot disengages from the artist's active workspace temporarily to reduce user heart rate.}
    \label{fig:painting}
\end{figure}

Eye-tracking technology has been used in artistic expression to uncover artists' intentions. A study used Tobii Pro Glasses 3 to examine Caravaggio's paintings~\cite{Eye}. This study aimed to determine whether Caravaggio deliberately considered the viewer's visual experience and how environmental factors might have influenced it. Motion capture systems like OptiTrack have enabled the creation of robot-specific movements in art, especially dance~\cite{Abe_2022}. This approach moves beyond simply mimicking human movements and focuses on the expressive potential of robotic motion. In this work, we aim to explore a combination of wearables to delete dynamic biometrics in a human artist, we will start by exploring heartbeat detection (see Fig.~\ref{fig:painting}).

\subsection{Robots that Paint}
Robots painting together with humans is not novel. The Robot Art Competition has resulted in many works created by robots~\cite{robotartcomp}. Another example is the Busker Robot which is able to paint watercolor images from reference~\cite{busker}. Many of these robot systems aim to replicate the input directly through the optimization of the pixels and brush strokes that the robot is able to create. However, Artist Sougwen Chung's work with developing robotic collaborators is an example of a more collaborative robotic system that responds to the brush strokes of the artist and is trained on a dataset of her previous works~\cite{sougwen}. E-David is another painting robot arm that mimics the manual painting process and can record and reproduce the same brush strokes~\cite{gulzow2020recent}.

The purpose of this project is to further explore the area in which artists and robots act as collaborators and to incorporate biometric data into our system. Our work builds on top of the FRIDA and CoFRIDA systems developed in~\cite{frida, s21237869}. They overcome limitations in robot painting such as the constraints of a robotic system/painting tools and the sim-2-real gap to create drawings from a trained text-to-image model using user input. CoFRIDA can also allow discrete turn-taking to promote collaboration between a user and the robot. However, we aim to further expand on the collaborative capabilities of the robot and show how such a tool could impact artists' creativity through modification of the FRIDA codebase and additional sensor data. 

\section{AURA System}


\begin{figure}
    \centering
    \includegraphics[width=\linewidth]{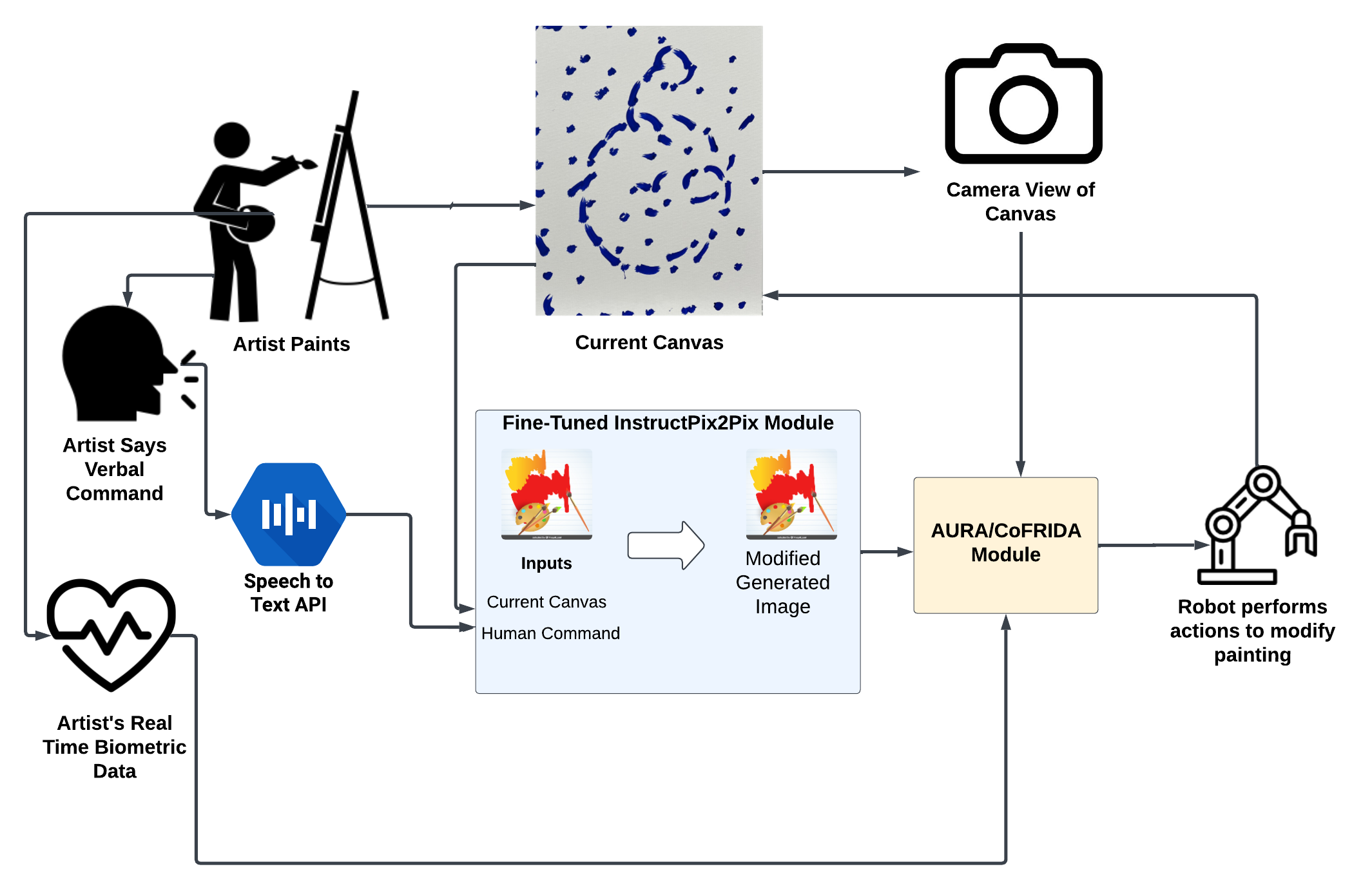}
    \caption{AURA System Setup. In addition to base CoFRIDA components, like the InstructPix2Pix image generation module, new inputs include user biometric data (heart rate), verbal commands, and additional camera views, allowing for seamless artist-robot interaction during the co-painting process.}
    \label{fig:enter-label}
\end{figure}

The robot is the main computational unit that determines what actions the robot should take, which include executing different strokes while painting, avoiding collision when a human is painting, reacting to human commands, etc. We intend to build a ROS 2 workspace around the existing FRIDA system, which currently determines a set of strokes/actions for the robot to paint a complete painting from a prompt. Specifically, we will still be using the determined robot actions generated from the pre-trained model using brush stroke data and existing artworks. However, instead of directly executing the motions, we will maneuver the robot based on the human position.

We use an off-the-shelf neural network to obtain the human joint positions in real time using a camera. The joint positions of the artist will then be passed into siMLPe, a motion prediction network ~\cite{simlpe}. This information will be published in our ROS 2 network. With the predicted motions, we will then use MoveIt 2 to plan a safe trajectory of the robot around the dynamic human obstacle. Also, we will modify the overall planning and the strokes based on biometric data such as the heart rate of the human or the motions of the human's paintbrush. All of the actions the robot can do, such as getting paint, responding to the biometric data, and painting strokes on the canvas will be managed by our state machine in ROS 2. Segmenting the painting and robot planning will also improve the collaboration of the robot, as we can replan the robot motion more often based on the artist's additions to the painting. This could also help the robot work in specific areas of the painting instead of all over the canvas, which could impede the artist's movements. 

A novel aspect of our work resides in that the robot takes in different types of data, including perception information, audio inputs, and biometric data. For the painting capabilities, the robot is constantly looking at the drawing being created and updating its steps accordingly based on the objective that is currently being followed. This is done using a DSLR camera. 
Audio input is used to translate the human artist's commands to shape the robot's movements. 

The verbal communication component of the system is built on top of AssemblyAI, a low-latency LLM-based speech-to-text transcription service, which allows user's verbal commands to be transcribed in real-time and relayed as an input to the InstructPix2Pix (image-generation) module.


A main component of the system is the biometric data, which enables the robot to understand the intentions of the artist. The EmotiBit (\url{https://www.emotibit.com/}) is a research-grade open-source wearable biometric sensor package that provides useful information that can guide the robot's current actions and future steps.

A question that is being investigated for this project is determining which biometric data streams are useful indicators of the human state and how a collaborative painting robot should interpret these data. Heart rate data is well understood to be linked with user arousal while performing tasks ~\cite{arousal} and can be used to inform robot behavior. Ideally, the interaction between the human and the robot should be fluid and minimally-invasive. The EmotiBit proves useful in this respect because of its small size and wide array of recorded data. For the purposes of this system, the EmotiBit wirelessly streams heart rate data which is then analyzed to inform the active state of the robot.

\section{Artist-Robot Interaction}

During our demo, we aim to showcase three different capabilities that our AURA system implements:  (1) the robot’s ability to take in heart rate data and adjust its behavior accordingly, and (2) 
the use of verbal commands from the human artist to instruct the robot.

\subsection{Painting Together Using Verbal Interactions}
\par\indent Since the human artist's hands are occupied with the task of painting, we incorporated a hands-free communication method for the robot using verbal commands. This approach ensures a smooth transition between tasks and allows the robot to receive quick feedback from the human artist. As detailed in the transcription section, we are using a large language model (LLM) to transcribe audio into input for the image generation model. Verbal commands are transcribed and then classified as either a ``painting command", which is then fed into the image generation model to be converted into painting actions, or a "direct command", which overrides painting commands and directly affects the robot's action outcome- commands like `Stop painting' or `Change colors'. 

\subsection{Heartbeat and Proximity}

We focused on human arousal as a determining factor for the robot’s proximity to the artist, drawing inspiration from the common human need for "space" in stressful situations. Higher heart rates and increases in heart rates are correlated with increased arousal, while lower, steady heart rates are correlated with decreased arousal. When the robot interferes with the human artist during painting, such as by making a mistake, the artist's heart rate increases. In our current system, if the heart rate exceeds a threshold defined by our linear regression model (i.e., the artist is classified as aroused), the robot retreats from the artist’s immediate workspace. To manage proximity, we divide the canvas into four quadrants: one represents the artist’s active workspace, the diagonal quadrant signals complete withdrawal, and the remaining two serve as intermediate zones when predictions fall near the threshold. As the artist’s heart rate lowers, indicating reduced arousal, the robot re-enters the collaborative painting zone. This creates a closed-loop interaction between human and robot. 

\section{Artist Impressions}
Working with a robot while painting can be challenging. There are several factors to consider, such as the robot not executing tasks exactly as the painter intends, obstructing the painter’s view and movement, and the robot’s limited mobility and artistic stylization. We observed that the artist had to maneuver around the robot arm and adapt to its presence rather than collaborating seamlessly. 
However, when these issues did arise, the closed-loop system allowed for some flexibility.
%
From an artistic perspective, we found that the robot sparked creativity. During the demonstration, the robot added yellow paint to a brown vase, inspiring the artist to introduce a cool tone to contrast the yellow and add dimension to the vase. Similarly, the geometric flowers painted by the robot inspired the artist to render a more detailed vase. Each artist brought their unique stylization to the collaboration, which was reflected in the final painting. 

\section{Future Work}
This system demonstrates a valuable opportunity to explore the various methods that humans and robots can interact in creative fields. As such, the system will act as a platform for continued development.

\subsection{Physically Moving the Robot In and Out of the Canvas}
The physical interaction between the human and the robot is intended to provide a direct and intuitive method for the artist to control the robot's painting location. If the robot is physically moved beyond the boundaries of the canvas, then it will cease painting, interpreting this action as a command to disengage. This behavior was intended to allow the artist to easily stop the robot's actions without needing to use verbal commands. However, if the robot is moved to a new location within the canvas, then the robot would interpret this as a signal to reposition and continue painting in the new location. This feature allows the artist to physically guide the robot's focus on specific areas of the canvas. The robot system should also be able to handle multiple input modes simultaneously, for example, receiving a verbal command after being physically repositioned within the canvas boundaries. Analyzing the robot's response to this combined input will offer valuable insights into how the system prioritizes and integrates different types of commands. Further investigation into multimodal command processing will be crucial for developing a more nuanced and responsive collaborative system.

\subsection{Additional Biometrics}
Incorporating multiple sensor types or streams could yield a more accurate prediction of the human state and, therefore, a more intuitive collaborative painting experience. The EmotiBit provides a host of biometric information about the user, so utilizing more of these data streams could allow for different characteristics of the painting experience to be observed and acted upon. One area of particular interest is mapping human emotions to the Russell’s Circumplex Model to gauge the artist's state in multiple dimensions. This additional information could foster a more responsive and adaptive collaboration between the artist and the robot. For instance, systems like CoFRIDA have successfully incorporated emotion into their generative image algorithms, showcasing the potential of emotion-aware collaborative systems. Additionally, there is merit to using the Tobii Glasses 3 to detect gaze in real-time as well as pupil dilation, but this has not yet been integrated into the system.

\subsection{Voice Integration}
To improve upon the system's voice integration, we are developing a machine-learning algorithm that does not require an explicit audio cue to activate. Instead, it uses a database of common commands to infer when to take action. This will enable the artist to communicate more freely with the robot. For example, the artist can issue commands such as ``Draw a dog'', ``Change colors'', and ``Stop''. They can also provide verbal feedback like ``Good job'' or `I did not like that''. This dynamic dialogue will foster an interaction that feels more akin to collaborating with another human artist.

\subsection{System Questions}
Given the complexity of our system, several open questions remain that we wish to address in future iterations:

\begin{itemize}
    \item Handling Conflicting Commands: How should the robot prioritize or resolve conflicting inputs from the artist?
    \item Streamlining Setup: Can the robot's setup and preparation time be minimized to improve usability?
    \item Adaptability: Can the robot learn and adjust its behavior based on the artist's preferences over time?
    \item Alignment with Artistic Vision: How can the robot's contributions better align with the artist's intent?
    \item Interactive Modifications: Can the artist alter the generative image or provide real-time adjustments while the robot is painting?
\end{itemize}


\section*{Acknowledgments}
\noindent We thank Peter Schaldenbrand and Jean Oh for the guidance to set-up the CoFrida system. This project is funded by DARPA Young Faculty Award (YFA) \#D24AP00323-00.

\bibliographystyle{abbrv}
\bibliography{appendix}

\end{document}